# Adversarially Perturbed Wavelet-based Morphed Face Generation

Kelsey O'Haire, Sobhan Soleymani, Baaria Chaudhary, Poorya Aghdaie,
Jeremy Dawson, and Nasser M. Nasrabadi, *Fellow, IEEE*
West Virginia University

*Abstract*— Morphing is the process of combining two or more subjects in an image in order to create a new identity which contains features of both individuals. Morphed images can fool Facial Recognition Systems (FRS) into falsely accepting multiple people, leading to failures in national security. As morphed image synthesis becomes easier, it is vital to expand the research community's available data to help combat this dilemma. In this paper, we explore combination of two methods for morphed image generation, those of geometric transformation (warping and blending to create morphed images) and photometric perturbation. We leverage both methods to generate high-quality adversarially perturbed morphs from the FERET, FRGC, and FRLL datasets. The final images retain high similarity to both input subjects while resulting in minimal artifacts in the visual domain. Images are synthesized by fusing the wavelet sub-bands from the two look-alike subjects, and then adversarially perturbed to create highly convincing imagery to deceive both humans and deep morph detectors.

## I. INTRODUCTION

Facial Recognition Systems (FRS) have become commonplace at border security crossings. With ever-increasing accuracy and speed, FRS are considered the premier method of obtaining biometrics. The International Civil Aviation Commission (ICAO) designated facial recognition as the required biometric in their electronic Machine-Readable Travel Document (eMRTD) because of the face modality's cultural acceptance, unobtrusive nature, and easy enrollment [1], [2]. If required, face data can be verified by a human, making it particularly attractive to border crossings where access to advanced verification technology may be limited [3]. While FRS are becoming a security necessity, they are vulnerable to attacks. The ICAO outlines the stages of a biometric system as: enrollment, template creation, identification, and verification. Bad actors can leverage vulnerabilities in the enrollment stage of the pipeline by submitting tampered identification images [4]. An enrollment system contains two steps, a detector for detecting tampered imagery, and a verifier which ensures that the image submitted is of the intended individual. We focus on fooling both the steps in the enrollment process.

Ferrara *et al.* [4] were the first to expose the dangers of morphed images being submitted for enrollment in an FRS. Morphed images are created by combining face images from two or more individuals creating a new ambiguous face which possess similarities between the bona fide identities. Using a morphed image, a passport can be shared between two or more individuals. If a bad actor is attempting to cross a border, they can create a morphed image with an individual who is similar looking in order to create a highly convincing morphed passport photo. The synthesized image would allow the bad actor to easily pass through the border using the passport of the look-alike individual. Synthesized morphs between two or more identities who naturally look alike create an ambiguous face, causing high false acceptance rates in detectors [3]. Commercial off-the-shelf systems (COTS) as well as human verifiers are vulnerable to these high-quality attacks [5]. As morphing technology becomes more accessible, anyone can create high-quality morphed images with little to no technical background. With no way of detecting morphed images with a high degree of confidence, national security is at risk.

We introduce a new method of morphing utilizing the Discrete Wavelet Transform (DWT). The input images are warped and then subsequently blended using their wavelet decomposed sub-bands and then reconstructed into the final morphed image. The authors in [6] have demonstrated that it is possible to detect morphed images using the high-frequency wavelet sub-bands. This work is the beginning stage of leveraging the spatial-frequency wavelet domain to create high-quality morphs. After morphing, a visually indistinguishable amount of adversarial perturbation is applied to further increase the difficulty of detecting the morphed images in our passport-system pipeline. A high-quality morph image will have no obvious signs of tampering and will show similarity to all individuals combined in the morphing process.

In this paper, we broadly classify malicious examples into two categories: geometric and photometric. We describe a geometric adversarial example as a transformation applied to a face resulting in the change of facial landmarks, such as warping. On the other hand, photometric adversarial examples are those described by Goodfellow *et al.* [7] [8] which include adding structurally significant noise to an image, disrupting a classifier's ability to discriminate a class for an image. At the time of this publication, we believe that no large-scale dataset has been generated using both face morphs and adversarial perturbation.

## II. RELATED WORK

Morphed images can be described in two different categories, landmark and GAN-based. The landmark-based morph image generation typically consists of a three-step pipeline: landmark detection, warping, and blending. Landmark morphs utilize critical points on a subject's face to warp the image. The landmark points of the two input subjects are averaged together to create common landmarks. The images are then warped towards these common landmarks and blended to create the final image. For more information, see Section III.A.

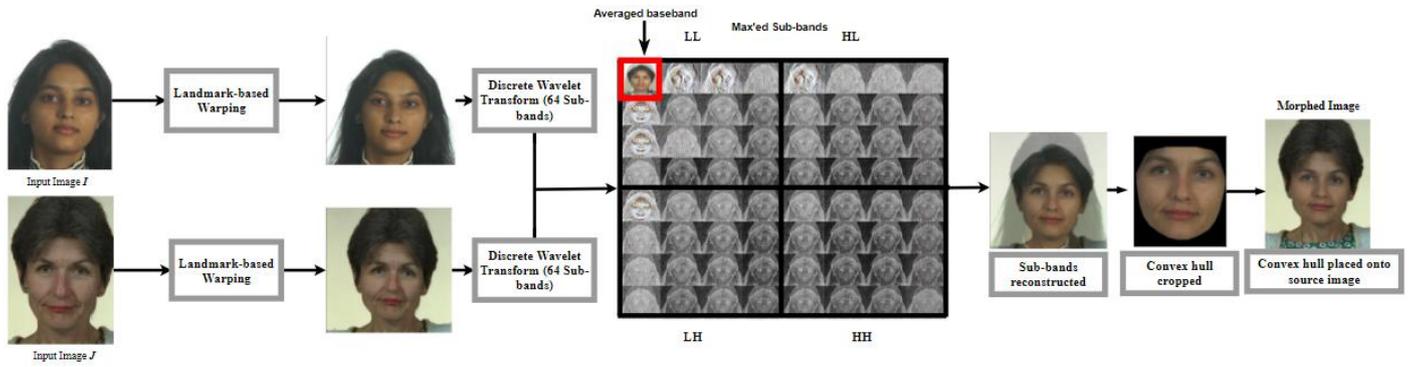

Fig. 1. Wavelet-based morphing pipeline. The input subjects are warped, and then wavelet decomposed into their respective 64 uniform sub-bands. The low-frequency basebands averaged together, and the remaining 63 sub-bands are max'ed together. The resulting sub-bands are then used to reconstruct the morph image, cropped, and placed on the input subject.

*A. Face Morphing*

In this work, we will focus on landmark-based morph generation. Ferrara *et al.* [4] morphed their images by hand using the open-source image editor GIMP. While the resulting images showed little artifacts, the pipeline was tedious and inconvenient to be scaled up for generation of large datasets. Since then, many open-source repositories have emerged, making it simple to generate large-scale datasets with ease. Sarkar *et al.* [9] generate three morphed datasets utilizing four popular morphing repositories: Facemorpher [10], OpenCV [11], WebMorph [12], and StyleGAN2 [13]. Facemorpher, WebMorph, and OpenCV are typical landmark-based repositories that rely on a combination of warping and splicing to generate images. The images are guaranteed to have visual similarity between both input individuals because features of the individuals are combined by averaging the input images together. While landmark-based morphing techniques are fast and effective, they tend to lead to artifacts in the final image, especially around the eye and background regions. StyleGAN2 is a Generative Adversarial Network (GAN) approach to face morphing where latent vectors of images are linearly combined, resulting in minimal artifacts and producing high-quality morphs [9], [13]. However, GAN-based approaches have issues retaining identity information after morphing, causing morphs to be more heavily weighted toward one subject than another, resulting in ineffective morphs [9].

*B. Adversarial Perturbation*

Adversarial perturbation is added to the morph images with the intention of fooling a morph detector into labeling the input as a bona fide class. Typically, the pixel values are constrained to an $L_\infty$ value which help to preserve the quality of the perturbed image. Adversarial perturbation should not be perceptually visible in the final image. Goodfellow *et al.* [7] introduce the fast gradient sign method (FGSM), which perturbs the input of the model based on the sign of the gradient for a target class. Liao *et al.* [14] utilized FGSM with a masking technique to perturb areas deemed as high importance using spatial information derived from multiple convolutional layers in a model. Hussain *et al.* [15] leverage adversarial perturbation for their work on adversarial deepfakes by perturbing frames of a video labeled as fake by a detector with the intention of all output frames being labeled as real.

### III. WAVELET-BASED MORPH GENERATION

*A. Landmark-based Wavelet Morphing*

We utilize a modified version of Facemorpher to morph our images [10]. Two identities are used for morphing: **I** and **J**. Input identities **I** and **J** are assumed to naturally look alike. **I** and **J**'s respective images $i$ and $j$ should be aligned. For our morphing pipeline, we utilize a landmark-based approach. 68-landmark points are found on input images, creating the 68-element long pixel-coordinates $\hat{\imath}, \hat{\jmath}$. Delaunay Triangles are utilized to create a mesh across the image, with the vertices of the mesh at $\hat{\imath}, \hat{\jmath}$. The $\hat{\imath}$ and $\hat{\jmath}$ are averaged together to create common landmarks, $\hat{k}$. An affine transform is used to map landmarked points from $\hat{\imath}, \hat{\jmath}$ to the $\hat{k}$, synthesizing $\hat{\imath}_{warped}$, $\hat{\jmath}_{warped}$. Bilinear interpolation is performed on the warped images to correct color values. At the end of the warping stage, the two images share common landmarks $\hat{k}$. After warping, $\hat{\imath}_{warped}$, $\hat{\jmath}_{warped}$ are decomposed into 64 sub-bands using a three-level wavelet decomposition. A vertical and a horizonal filter are applied to the warped images, creating the Low-Low, Low-High, High-Low, and High-High sub-bands. As presented in Figure 1, the low frequency baseband after three-level wavelet decomposition of the $\hat{\imath}_{warped}$, and $\hat{\jmath}_{warped}$ are averaged together. This sub-band is selected because it represents most of the shared information from the original subjects. The remaining 63 sub-bands are combined using the maximum-coefficient at every location in the sub-bands to capture the most significant information from each subject. Once the two input images are wavelet morphed, the convex hull of the morphed image is spliced onto the background of the source and destination images.

*B. Adversarial Perturbation for Morphed Image Generation*

While the wavelet-based and standard morphs may be able to fool a person, a trained deep-learning based morph-detector is still able to detect morphed images in the enrollment stage. Therefore, we fine-tune an Inception-ResNET v1 model [16] pretrained on VGGFace2 [17] to detect morphed images based on the work from the authors of [18].

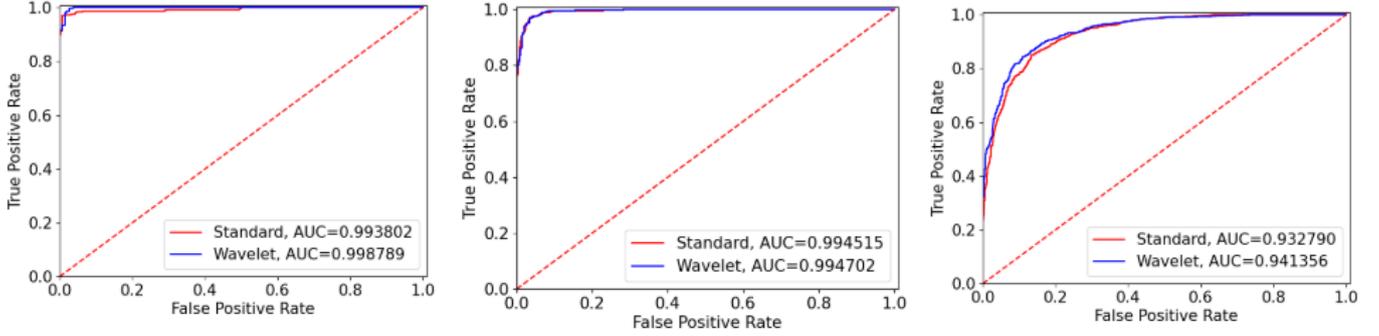

Fig. 2. FaceNet as a verifier: ROC curves for the (left) FRLL, (middle) FERET, and (right) FRGC datasets.

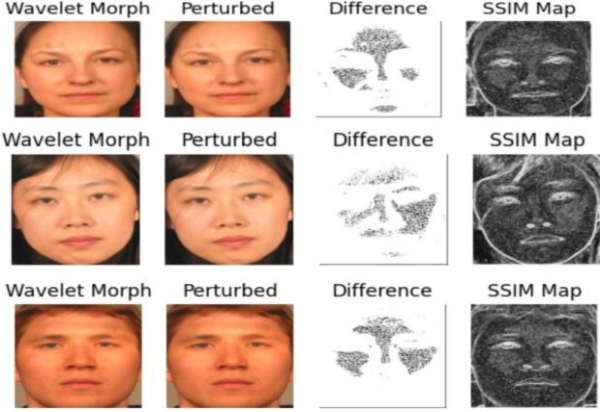

Fig. 3. Wavelet morphs with respective perturbed images from the FRGC dataset. The Difference column represents the min-max normalized absolute difference between the wavelet morph and their respective perturbed wavelet morph. The SSIM Map column represents the areas of high similarity between the input images. In the difference and SSIM images, white and black represent smaller and higher values, respectively.

The model was trained on 4,000 morphed images. The morph detector was able to detect morphs with near perfect accuracy. To further make the morphs harder to detect, adversarial perturbation is applied to the wavelet-based morphed images. Using our morph detector, images are perturbed using the Basic Iterative Method (BIM) [8]. FGSM perturbs an image based on the gradient with every iteration of backpropagation. An $L_\infty$ is used as a maximum allowed pixel difference constraint. BIM is a derivation of FGSM, where a constant step-size is utilized for every applied perturbation. BIM is formulated as:

$$X_{N+1}^{adv} = Clip_{X,\epsilon}\{X_N^{adv} + \beta \text{sign}(\nabla_X L_{adv})\}, \quad (1)$$

where $X_0^{adv} = X$ is the original morphed image and $L_{adv}$ consists of cross-entropy loss as well as the Total Variation (TV) smoothing loss:

$$L_{adv} = J(X_N^{adv}, y_{true}) - \lambda \, TV(X_N^{adv}), \quad (2)$$

where $J$ is the cross-entropy cost function between the adversarial image and the target class, $\beta$ is the perturbation step size and $\epsilon$ is the $L_\infty$ constraint on the pixel values [8]. $Clip_{X,\epsilon}$ confirms that the pixel values are within $\epsilon$ $L_\infty$-norm distance from the original sample. We also clip the adversarial example

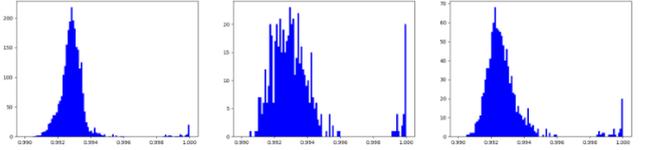

Fig. 4. SSIM distributions between wavelet morphs and their respective perturbed wavelet morphs for (left) FRLL, (middle) FERET, and (left) FRGC datasets.

at each iteration to make sure that all pixel values reside within the valid input range.

In addition, $\lambda = 0.1$ is the regularization parameter. To further help the visual quality of the image, TV smoothing was applied to the added perturbation to remove any visible artifacts in the adversarial image [19]:

$$TV(X_N^{adv}) = \sum_{i,j} ((r_N[i,j] - r_N[i+1,j])^2 - (r_N[i,j] - r_N[i,j+1])^2)^{\frac{1}{2}}, \quad (3)$$

where $r_N[i,j]$ is a pixel in the perturbation image $r_N = X_N^{adv} - X$ [20]. After perturbation, the images are expected to maintain their perceptual quality.

## IV. RESULTS

We utilize the FERET, FRLL, and FRGC datasets for our morphs [21] [22] [23]. The datasets contain image sizes of 413×531 for FRLL, 1704×2272 for FRGC, and 512×768 for FERET. Each dataset depicts passport style images with a neutral face looking into the camera under ideal lighting conditions. In total, FERET contains 1,199 different identities, FRLL contains 102 identities, and we used a subset of FRGC which contains 765 identities. Our morphing methodology follows that of Facemorpher [10]. We use morphed images from [9] for comparison to our work and we refer to these image as standard morphs for the rest of this paper because they use a typical morphing pipeline consisting of an alpha-blending step used to combine warped images. For pairing our morphs, we use the protocols originally created by Neubert *et al.*'s AMSL dataset for FRLL [24] and Scherhag *et al.*'s protocol for FERET and FRGC [25]. In addition, landmarked-based wavelet morphing images described in Section III.A are referred to as wavelet morphs. In total, 529 FERET-based, 1,222 FRLL-based, and 924 FRGC-based wavelet morphs are generated.

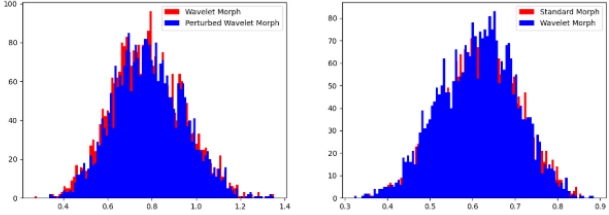

Fig. 5. (left) FaceNet match score ($L_2$) distribution and (right) SSIM score comparison between input subjects and their respective morphs.

### A. Similarity Comparison

Morphed images share features from both input subjects; therefore, a quantitative measure of perceived similarity is needed for comparison. We use two different metrics, a FaceNet match score [26] and the Structural Similarity metric (SSIM) [27]. Both metrics are selected because they represent a perceived similarity rather than a direct pixel-comparison to their input images. FaceNet is leveraged to quantify look-alikes from a deep learning verifier, while SSIM uses classical techniques to quantify perceptual similarity. To minimize extraneous information from the morphed images, the convex hull region of the morph is extracted for the comparison.

FaceNet uses a deep convolutional network architecture to create a compact feature embedding of its input. FaceNet is trained using triplet loss, where the Euclidean distance ($L_2$) for embeddings of the same identity are positive examples and differing identities are considered negative examples [26]. Therefore, there is a correlation between the $L_2$ distance of feature embeddings and perceived similarity. SSIM is a quality metric used to mimic similarity of two images as perceived by the human eye. SSIM is calculated using a combination of three independent comparisons: luminance, structure, and contrast [27]. Visible artifacts in an image decreases the SSIM.

Figure 5 shows the distributions of the $L_2$ distance between FaceNet embeddings and SSIM comparison from the standard morphing technique described in [4] and our wavelet morphing technique. For both distributions, every morph has two separate comparison values, one comparison to subject **I**, and one to subject **J**. For our FaceNet comparison, a smaller $L_2$ value represents a stronger look-alike. Inversely, a larger SSIM value represents a stronger look-alike. Both distributions show that our wavelet-based morphing technique is as effective at creating morphs that look like their input subjects as [9], while retaining the image quality. The standard and wavelet-based morphs share the same mean of their SSIM distributions (SSIM of 0.61), resulting in no difference between visual quality of the standard and wavelet-based morphs. On the other hand, for the perturbed images, SSIM is found between the wavelet morphed image and the perturbed wavelet morphed images. If the images are too heavily perturbed, they exhibit signs of degradation. The SSIM score for all datasets show that every perturbed image has an SSIM of above 0.99, indicating that all images are perceived

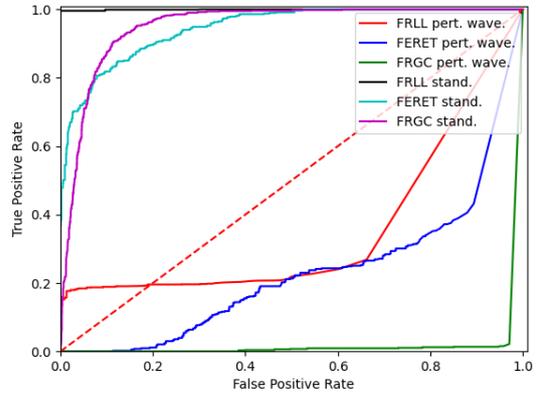

Fig. 6. ROC curves for trained morph detector.

to be indistinguishable to the wavelet-based morph. Figure 4 shows the distribution of SSIM values.

### B. White-box Detector and Verification

As presented in Figure 2, FaceNet can differentiate between morph and genuine images when compared to a reference photo for both our wavelet morph and Sarkar *et al.* [9]. The wavelet-based perturbed and standard images are tested on the trained morph detector and the results are shown in Figure 6. In Equation (1), we use $\beta = 6$ and $\epsilon = 2$ for perturbation. Image perturbations take approximately 2 seconds per image. The AUC for FRGC is 67%, FERET is 24%, and FRLL is 2%. The results show that the perturbed images are being erroneously classified as bona fide images at an alarming rate. Our perturbed wavelet morphs would bypass a morph detector in the passport pipeline with a high degree of success.

To determine the morph's effectiveness on the verification stage, we utilize a pretrained FaceNet model as a verifier [26]. A reference image of a subject is compared to a second genuine image of the subject to create a positive comparison, and a negative comparison is made between the reference image and its respective morph. FaceNet ROC curves are plotted for each of the datasets as shown in Fig. 2. The true positive score signifies a morph correctly labeled as a morph. FaceNet discerns the wavelet morphs at a nearly identical rate as the standard morphs. The verification stage is the most likely point in the pipeline for the morph to be detected. Verification is a difficult problem for face morphing because the morph image must contain features from both input subjects, making it difficult for the resulting morph to appear more similar to a reference image than to a bona fide image.

## V. CONCLUSION

In this paper, we provided the prospect of morphing in the spatial frequency wavelet domain. We showed that our wavelet-based morphs are as convincing as morphs generated in prior works, while introducing a new morphing methodology. By adding adversarial perturbation, the wavelet-based morphs are nearly impossible to detect by humans and deep learning-based detectors. In the future, more sophisticated methods of sub-band selection can be used to generate morphs in the spatial-frequency domain, creating morphs that are more difficult to detect.